\documentclass{article}

\usepackage{arxiv}

\usepackage[utf8]{inputenc} 
\usepackage[T1]{fontenc}    
\usepackage{hyperref}       
\usepackage{url}            
\usepackage{booktabs}       
\usepackage{amsfonts}       
\usepackage{nicefrac}       
\usepackage{microtype}      
\usepackage{cleveref}       
\usepackage{lipsum}         
\usepackage{graphicx}
\usepackage{natbib}
\usepackage{doi}
\usepackage{todonotes}

\title{Technical Language Processing for Telecommunications Specifications}

\newif\ifuniqueAffiliation
\uniqueAffiliationtrue

\ifuniqueAffiliation 
\author{Felipe A. Rodriguez Y.\thanks{The author works as Entity Specifications Engineer at Nokia Finland. With several years of experience in the RAN, he is currently exploring Generative AI use cases associated to technical specifications.} \\
	L3 A\&S\\
	Nokia Networks Oy\\
	Espoo, Uusimaa 02610 \\
	\texttt{felipe\textunderscore a.rodriguez\textunderscore yaguache@nokia.com} \\
	 \\
}
\else
\usepackage{authblk}

\newbox{\orcid}\sbox{\orcid}{\includegraphics[scale=0.06]{orcid.pdf}} 
\author[1]{%
	\href{https://orcid.org/0000-0000-0000-0000}{\usebox{\orcid}\hspace{1mm}David S.~Hippocampus\thanks{\texttt{hippo@cs.cranberry-lemon.edu}}}%
}
\author[1,2]{%
	\href{https://orcid.org/0000-0000-0000-0000}{\usebox{\orcid}\hspace{1mm}Elias D.~Striatum\thanks{\texttt{stariate@ee.mount-sheikh.edu}}}%
}
\affil[1]{Department of Computer Science, Cranberry-Lemon University, Pittsburgh, PA 15213}
\affil[2]{Department of Electrical Engineering, Mount-Sheikh University, Santa Narimana, Levand}
\fi

\renewcommand{\shorttitle}{\textit{arXiv} Template}

\hypersetup{
pdftitle={Technical Language Processing for Telecommunications Specifications},
pdfsubject={q-bio.NC, q-bio.QM},
pdfauthor={David S.~Hippocampus, Elias D.~Striatum},
pdfkeywords={First keyword, Second keyword, More},
}

\usepackage{caption} 
\begin{document}
\maketitle

\begin{abstract}
Large Language Models (LLMs) are continuously being applied in a more diverse set of contexts. At their current state, however, even state-of-the-art LLMs such as Generative Pre-Trained Transformer 4 (GTP-4) have challenges when extracting information from real-world technical documentation without a heavy preprocessing.  One such area with real-world technical documentation is telecommunications engineering, which could greatly benefit from domain-specific LLMs. The unique format and overall structure of telecommunications internal specifications differs greatly from standard English and thus it is evident that the application of out-of-the-box Natural Language Processing (NLP) tools is not a viable option. In this article, we outline the limitations of out-of-the-box NLP tools for processing technical information generated by telecommunications experts, and expand the concept of Technical Language Processing (TLP) to the telecommunication domain. Additionally, we explore the effect of domain-specific LLMs in the work of Specification Engineers, emphasizing the potential benefits of adopting domain-specific LLMs to speed up the training of experts in different telecommunications fields.
\end{abstract}

\keywords{Telecommunications\and Generative AI \and Natural Language Processing \and Technical Language Processing \and Technical Specifications}

\section{Introduction}

Technical specifications are documents containing requirements that need to be satisfied by a product. In the telecommunications field, these often refer to widely known open standards like 3GPP or O-RAN. However, although these standards serve as high-level guidelines for system architecture and implementation, telecommunications equipment vendors write their own extensive internal technical specifications containing careful descriptions of product behavior. This internal documentation has a complex format, contains proprietary data, and is the main source of knowledge during the telecommunications software development phase. Studying such specifications and abstracting as much information as possible from them in a fast and reliable way is one of the key challenges faced by engineers in the telecommunications industry.

The rapid advancement of Natural Language Processing (NLP) is evidenced in a wide variety of application domains, driven by the improvement of language models capable of extracting information from a determined context, and the recent emergence of Large Language Models (LLMs). One example of an LLM is the Generative Pre-Trained Transformer 4 (GPT-4) \cite{achiam2023}, which is the most capable LLM available at the moment of writing this article. GPT-4 has performs well in a range of tasks from general question answering to code generation \cite{joublin2023}.

Nevertheless, there is still plenty of room for improvement for the telecommunications industry trying to automate the extraction of information from internal documentation. Although showing impressive results for applications working with generic knowledge obtained from standard datasets, it is clear that state-of-the-art language models fail in generalizing the knowledge obtained during the training phase when exposed to domain-specific tasks. The traditional approach for deal with domain-specific tasks is to fine-tune the models built with data from a resource-rich domain using data from a low-resource technical domain. There are two significant issues that arise when following this fine-tuning approach. First, the amount of domain-specific data is in reality not low but is protected by intellectual property rights. Second, the format of internal documents differ greatly not only from that of traditional English language but also between the areas where these documents were written.

\begin{figure*}[t]
  \centering
  \includegraphics[width=\textwidth]{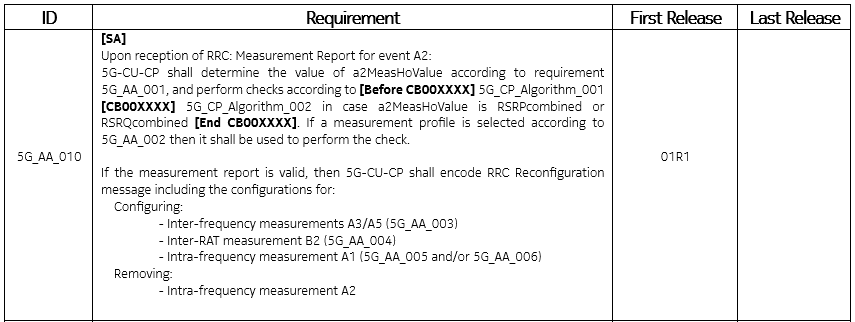}
  \caption{An example of internal technical specification requirement.}
\end{figure*}

In this article, we explore the use of Technical Language Processing (TLP) in Telecommunications,  looking for how to maximize information extraction from the internal technical specifications. Our contributions to the field are as follows: 

\begin{enumerate}
    \item Characterization of internal specifications.
    \item Identification of the limitations faced by out-of-the-box NLP tools when dealing with internal specifications data.
    \item Conceptualization of TLP in the telecommunications industry.
    \item Analysis of the challenges found in internal specifications that hinder the application of TLP.
\end{enumerate}

This article is organized as follows. Subsequently, in Section~\ref{sec:related}, we outline related works focusing on leveraging LLMs on the internal documentation in the telecommunications industry. In Section~\ref{sec:internal_specifications}, we outline internal documentation, discussing their role and format. Then, in Section~\ref{sec:nlp_limitations}, we discuss the limitations of NLP on technical data, and highlight the need for TLP technical specifications. Section~\ref{sec:tlp_in_telecom} builds on an existing TLP concept, highlighting the unique characteristics of technical specification in the telecommunications industry, and Section~\ref{sec:gen_ai_friendly} outlines a set of recommendations to achieve Generative AI-friendly internal specifications. Finally, in Section~\ref{sec:conclusion}, we explore future research directions.

\section{Related Work}
\label{sec:related}

To the best of our knowledge, this is the first work that investigates the role of LLMs in the telecommunications industry focusing on knowledge extraction from internal documentation. However, some authors have previously focus on the need of TLP for handling complex technical text. The authors in \cite{Dima2021} investigated the adaption of natural language for the processing of technical text belonging to industrial maintenance orders. More specifically, the authors define the role of TLP as an iterative approach that encourages human experts’ knowledge injection. In \cite{Brundage2021}, the authors analyze the issues of NLP tools when dealing with technical text coming from maintenance orders, and how it affects knowledge extraction. In \cite{Sarica2021} an analysis of the role of stopwords in language processing is done and a technical stopwords list that could be used for text filtering in engineering applications is created. However, the list is extracted from the titles and abstracts of utility patents from USPTO, thus not focusing solely on Telecommunications. \cite{Nandyala2021} offers different approaches such as word and sentence similarity, as well as word clustering for evaluating the representation of words in industrial scenarios. In \cite{karim2023} the authors pretrained an LLM model using a custom large dataset containing information from telecommunication open standards, as well and available online sources. Results showed that pretraining a model using the customized datasets and further finetune these models for specific tasks provide more insightful answers than using out-of-the-box models.

\section{Internal Specifications}
\label{sec:internal_specifications}

Telecommunications equipment vendors can nowadays be classified as software companies, as a greater part of their business is to provide periodical and on-demand equipment software updates to their customers. In this manner, it is imperative for equipment vendors to provide high-quality software in a fast and efficient way. As new telecommunications use cases emerge and customers become more heterogeneous, the demand for customized network behavior is constantly increasing. This is where internal technical specifications are needed, as they maintain a clear description of all the implemented functionalities in different levels of detail. These documents are the implementation guidelines for software developers working on a specific project, who before writing code need to carefully understand the designed functionality.

\subsection{Writing of Internal Specifications} 

\begin{figure*}[t]
  \centering
  \includegraphics[width=\textwidth]{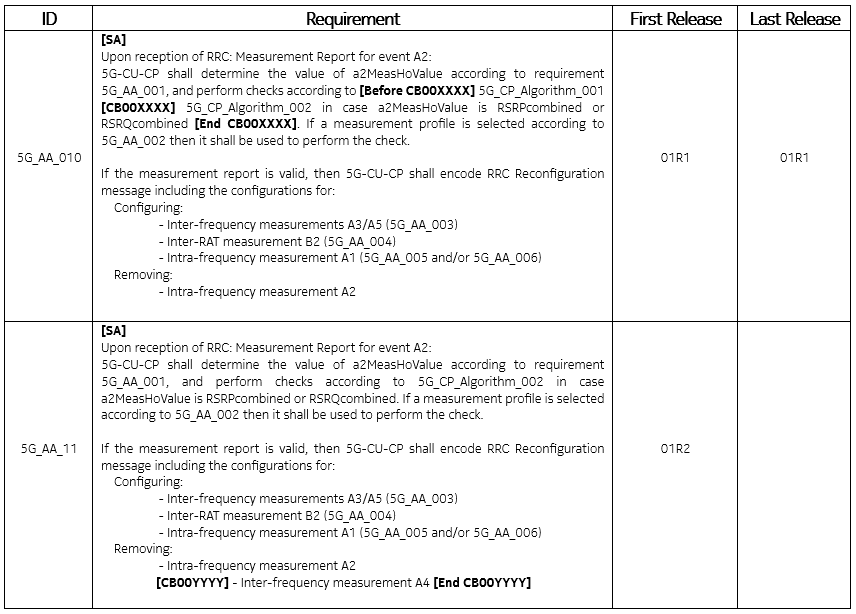}
  \caption{An example of internal technical specification requirement with different behavior for two different releases.}
\end{figure*}

Internal specification is written by engineers during the development of a new telecommunication network functionality and is a quite data-driven process. During the specification writing phase, an intensive study of the functionality being implemented is necessary, as well as an analysis of the dependencies with previous developments i.e., the ways this new development affects or does not affect an existing behavior. The size of the internal specification depends on the complexity of the  functionality being implemented, and can result in a couple of short requirements in a single chapter within the same document, to dozens of medium or large-size requirements distributed in different modules. 

Typical internal specification data is different from standardized documentation, and contains (i) messy raw text and/or tables; (ii) complex internal data formats; (iii) inconsistent, ambiguous, grammatically incorrect, and incomplete text; (iv) duplicated data within the same document as well as between different documents. These data issues affect not only developers in charge of implementing the described functionalities, but also the specification engineers designing such functionalities and writing the documentation as they need to understand previous behavior in order to determine the changes to be done.

\subsection{Characteristics of Technical Specifications}

An example of internal specification requirement can be seen in Fig.1. Due to simplicity, the requirement examples followed in this work do not include all the existing properties and focus on the most important ones. Three characteristics are clearly visible: a unique ID, the requirement content, and the release in which the requirement is valid. In order to properly interpret the content of a requirement we need to focus on two out of the three components, its content and version. Within the content of the analyzed requirement, we can see that the are three bold tags {\bfseries [Before CB00XXXX]} specifying the behavior that has been replaced by the new development, {\bfseries [CB00XXXX]} introducing the new behavior, and {\bfseries [End CB00XXXX]} marking the end of the behavior modifications. Additionally, the tag {\bfseries [SA]} indicates the deployment type for which this requirement is valid, in this case standalone.

The release is a code useful for tracking the software changes, successfully associating certain behavior with a specific software version(s). More specifically, {\bfseries First Release} corresponds to the release where the requirement was introduced. On the other hand, {\bfseries Last Release} means the very last release for which the requirement is valid. In most cases, a functionality will be altered from one release to the next due to new developments. Fig.2 shows the same requirement but now it has two versions, 01R1 and 01R2. Also, notice that the previously explained tags of feature {\bfseries CB00XXXX} have now been deleted but the changes introduced by this development are maintained. Additionally, the new release 01R2 marks the point when the new behavior introduced by development {\bfseries CB00YYYY} starts to be valid, and the corresponding tags have been added.

\section{Limitations of NLP on Technical Specifications}
\label{sec:nlp_limitations}

Large Language Models (LLMs) have demonstrated enhanced capabilities far beyond those of Natural Language Processing (NLP) in several fields \cite{Kim2024}, however, they are still underutilized in the Telecommunications domain. Although research is steering towards the idea of a Telecommunications-capable LLM \cite{karapantelakis2024}, state-of-the-art research is still in its infancy and not capable to efficiently grasp knowledge from complex internal specifications. LLMs such as Generative Pre-Trained Transformer 4 (GPT-4) \cite{achiam2023} are trained on a large text corpora written in standard English language, which deviates from that used in technical requirements.

\subsection{Preprocessing Techniques}
Technical specifications pose a challenge to open source, state-of-the-art preprocessing techniques. One of these preprocessing stages is text cleaning \cite{egger2022}, which is ussually carried out by performing case lowering, non-word and non-whitespace removal based on a vocabulary, as well as digit removal from the data. While case lowering might not present any significant issues, the presence of highly technical words or syllables might hinder the removal of non-words since these would not be included in a general vocabulary. In the same manner, digit removal might not be recommendable as digits usually represent useful information in technical fields, for example an ID.
Another stage is word tokenization \cite{song2021}, in which text is split into words. During this stage the major issue is that technical text will not be appropriately split \cite{chai2023}, this occurs due to the tokenization rules not considering the unique content within the technical specification, for example parameter names or requirement IDs. Different tokenization tools will split technical text in different ways, thus producing a varying number of tokens for the exact same text.

\subsection{Context Building and Information Abstraction}
Even though LLMs are evolving towards supporting increased context lengths, long contexts created by using specification requirements cannot be efficiently used by an LLM to construct the answer to a query as demonstrated by \cite{liu2023}. Also, different from standard English and as explained in the previous section, requirements are written in a format that complicates information abstraction. As shown in Fig.1, when compared to the corpora of text used during pretraining \cite{liu2024}, an specification requirement is short, and contains many abbreviations or acronyms that are either specific to the telecommunications domain or the company the document belongs to. Words can as well be misspelled, contain typos, or are simply omitted.

\subsection{Need for TLP in Technical Specifications}

The main use case of NLP in telecommunications is text preprocessing prior to the extraction of information. As it has been already analyzed, using out-of-the-box NLP tools with technical specifications will not produce optimal results. NLP tools should not be completely avoided, but rather used for a task they are appropriate to. Text analysis and generation might be severely affected when using out-of-the-box tools for NLP that generalize the preprocessing techniques and are not capable to handle the uniqueness of internal technical specifications. 
Thus, there is a need to advance TLP for the telecommunications domain, and to further adapt it to consider the different information resources available within telecommunications equipment vendors.

\section{TLP in Telecommunications Industry}
\label{sec:tlp_in_telecom}

Perhaps the greatest innovation in TLP is the consideration of a human-in-the-middle acting as a source of knowledge. The concept introduced by \cite{Brundage2021} is valid for open-source technical text in mechanical engineering, when talking about the telecommunications field we need to further develop it considering the needs of the industry. One of the main objectives for the application of TLP in telecommunications is to extract knowledge from internal documentation, and as already explained in the previous chapter these documents usually have a proprietary format and contain sensitive information.

\begin{figure*}[t]
  \centering
  \includegraphics[width=\textwidth]{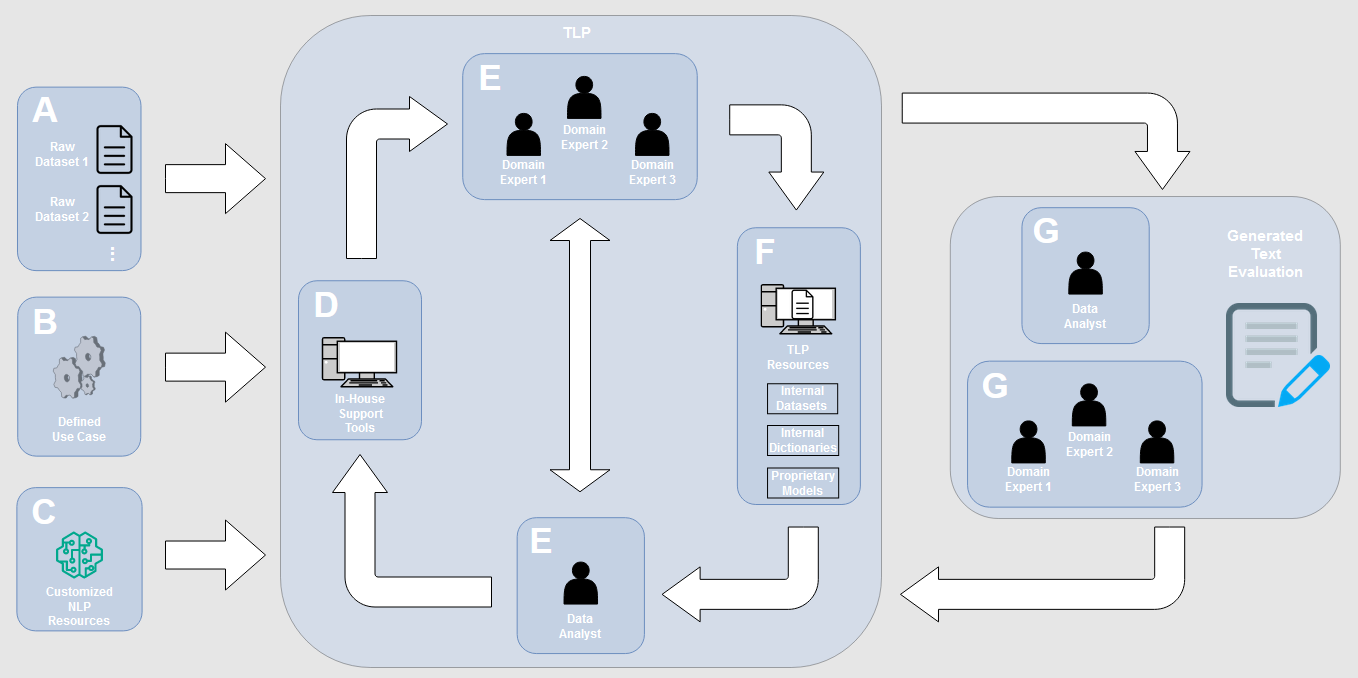}
  \caption{Diagram of the conceptual application of TLP to Telecommunications Internal Specifications.}
\end{figure*}

Fig3. shows the refined concept of TLP for its application on knowledge abstraction from internal documentation following a format as that presented in section 3, where:

\begin{itemize}
    \item[A] Raw data cannot be extracted in a single file, as internal specifications tend to be extremely long. Considering the requirement format explained in this work, one option is separating technical data by release. In this way, duplication is avoided and the resulting dataset is smaller.
    \item[B] Use case definition involves the selection of data sources. This step is important as document format and content vary between organizations. Furthermore, the use case definition shall set a scope based on the information that is possible to extract from the data sources.
    \item[C] The use of current out-of-the-box tools for language processing does not have the desired result when applied to data coming from internal documentation. Thus, in-house tools, often derived from open-source solutions need to be developed.
    \item[D] In-house support tools do not only help domain experts with their tasks, but also allow for fast information visualization and data extraction. Tools shall aim at improving overall working efficiency. 
    \item[E] Telecommunications is a quite broad field. An expert from a specific domain will not have a strong background in a different domain, and might not have any ML-related understanding. Similarly, data analysts usually do not have any telecommunications knowledge. Direct collaboration between domain experts and data analysts is vital for the implementation of TLP.
    \item[F] Internal TLP resources are iteratively built and maintained with the help of domain experts. In-house resources such as manually-built technical datasets are essential to fully exploit the potential of TLP. Furthermore, these internal resources could be reused for a wide variety of use cases based on their definitions. 
    \item[G] Analysis of the generated text is used to further improve the quality of TLP resources, as domain experts can provide feedback on the content and format of the text. However, the review process is quite tedious for the domain experts and new metrics to measure the technical quality of generated text is therefore necessary.
\end{itemize}

\subsection{Use Case Definitions}
Often overlooked in industrial environments. Properly defining a use case allows for the identification of the data to be used, the availability and format of this data, as well as the limitations of the use case. Technical queries expect specific answers, based on the introduced specification format, it is necessary to ensure that the following mapping of information is available to the LLM in order for it to efficiently abstract knowledge from the context:
\begin{itemize}
\item Mapping a procedure behavior with a release: in a development environment, telecommunications procedures are subject to minor modifications between releases due to new software developments. When extracting information from internal documents, this mapping will allow answering queries in the form \textit{“how does procedure X behaves in release Y?”}.
\item Contrast the behavior of a procedure between releases: similar to the previous mapping. With telecommunications procedures being affected by one or more changes per release, it is necessary to compare the behavior of certain procedure between two different releases. This mapping would allow to extract information for queries in the form \textit{“what is the difference in the behavior of procedure X between release Y and release Z?”}.
\item Mapping changes in a procedure behavior due to new developments: similar to the first mapping in the way that new developments modify the behavior of existing procedures. However, it is also necessary to identify what were the developments that modified such procedures. This mapping would allow answering queries in the form \textit{“how was procedure X modified by development Y?”}. 
\item Mapping a procedure with a requirement ID: since every requirement ID is unique, it would be useful to associate requirements with the procedure they describe. This mapping will allow to answer queries in the form \textit{“what requirements are related to procedure X?”}.
\item Mapping behavior with deployment type: a certain procedure may differ greatly between standalone (SA) and non-standalone (NSA) deployments. Thus, an association between a behavior and the type of deployment where it is valid is necessary. This mapping allows to answer queries in the form \textit{“how does procedure X behave in SA/NSA?”}.
\end{itemize}

\begin{table*}
  \caption{Analysis of Internal Specification Challenges}
  \label{tab:Challenges_Spec}
  \begin{tabular}{p{3.0cm} p{4.9cm} p{2cm} p{4.9cm}}
    Challenge & Description & Severity & Actions To Mitigate \\
    \midrule
    
    Duplication
    & Information copied from one requirement to another within the same document with slight modifications. And includes the case when information from an existing requirement is reworded, and added as a new requirement in a different module.
    & High 
    & Properly define the document or section where a requirement ought to be written. Also, specification requirements shall be written in a modular manner, this will reduce the need to duplicate content. \\ \hline
    
    Requirement length
    & Non-modular requirements covering several procedures, thus heavily duplicating information while adding only minor changes. The inclusion of different deployment behavior in a single requirement is as well part of this issue. Furthermore, long requirements tend to be unordered and deployment behavior alternates between NSA and SA.
    & High
    & Properly define the procedure to be written in a requirement. It should be considered that functionalities whose behavior vary depending on deployment type could be specified in separate requirements, one per deployment type.
    Overall, requirements shall be straight and to the point.
    \\ \hline
    
    Lack of standardization 
    & Different names are used for the same procedures or messages. In the same manner, deployment, feature, or requirement tagging is done in different ways, sometimes within the same requirement.
    & High 
    & Names of procedures and messages shall be internally standardized, the use of technical jargon shall be avoided. Naming conventions, as well as tagging conventions shall be enforced and effectively communicated. 
    \\ \hline

    Complex grammatics and semantics
    & Poor language skills result in typos, ambiguity, and reduce overall understanding. On top of this, format of requirement IDs, and parameter names are nothing like regular vocabulary.
    &Medium 
    & Final version of a requirement should be grammatically and semantically checked before being added to the internal documentation.    
    \\ \hline

    Disperse Information 
    & Similar or related topics are not grouped within a common section, and thus information might be scattered at the beginning, in the middle, and at the end of a document.
    & Low 
    & Requirements belonging to related procedures should be added to a common section.  
    \\
    
  \bottomrule
\end{tabular}
\end{table*}

\subsection{Data representation}
It is necessary that raw datasets possess the right data representation in order to achieve an efficient information extraction. Thus, a light formatting is necessary during data extraction with the following purposes: first, to avoid information duplication due to the existence of multiple releases; second, eliminating those requirements that do not contribute significant information, such as headers; and third, to reduce the size of the raw dataset. In this manner, a single raw dataset will contain requirements belonging to a single release avoiding any ambiguous or contradictory information between releases.

\subsection{Internal Datasets and Dictionaries}
Besides creating field-specific datasets containing open-source technical knowledge, there is need for internal datasets to be available as in contrast with generic datasets they will contain vendor-specific technical information, terminology, and descriptions. These resources will help to successfully develop TLP, as models further trained with internal datasets are capable of better information abstraction.

Furthermore, internal dictionaries are needed to achieve complex relationships within the requirements text. As an example, telecommunications procedure names can be written in many different ways, for example, "A2 measurement",  "A2 measurement for Handover", or "A2 measurement for the activation of Inter-frequency measurements" refer to the same procedure but are not consistent from the lexical point of view. 

\subsection{Proprietary Models}
The specification format alongside query complexity means it is not possible to achieve reliable answers based solely on building a long context, or deliberately fine-tuning a model using the internal documentation. Aforementioned TLP resources can be used to apply prompt engineering, train a customized model, or set of models that adapt to the defined use cases.

\section{Preparing Internal Specifications for the Deployment of Generative AI}
\label{sec:gen_ai_friendly}

The requirements introduced in this work are ideal, technical specifications are more complex and will usually include other properties that for the sake of simplicity will not be discussed in this work. However, there are still many potential improvements that can facilitate the deployment of generative AI solutions using internal documentation as a source of knowledge. Below we analyze three challenges found in internal documentation and we give recommendations on how to make more GenAI-friendly internal specifications.

\begin{itemize}
\item Content duplication: Requirements might duplicate content either within the same module, or between documents. Problems arise as the duplications tend not to be word-by-word copies but rather slight modifications of requirements, many times adding ambiguity. This severely affects the performance of any tool based on GenAI, as no model will be capable of identifying these kinds of changes. As an example, a parameter called “activateMeasurementSA” is mentioned in a requirement, that is linked to another requirement in a different module where the parameter was originally named “activateMeasurement”. It is necessary to avoid the copying or rephrasing of already existing behavior at all costs. 
\item Requirement length: Long and verbose requirements are not necessarily better than short and straightforward documentation. Additionally, excessively long requirements are problematic since they tend to include multiple procedures, resulting in a non-modular specification. Furthermore, this abuse of verbosity will ultimately end in a severe duplication of information, since many long requirements will often include steps that have already been defined in previous documentation. Overall, a lack of simplification affects the tokenization process as it leads to an increase in the token count without adding any extra information. Ensuring the atomicity of requirements is vital for improving information extraction in a multi-document environment. 
\item Lack of standardization: In internal documentation, the names of procedures or messages might not be standardized, meaning several different names will describe the same behavior. At the same time, the use of vendor-specific terminology further deepens this issue due to the lack of naming conventions. Going further, a lack of standardization in either deployment or development tagging complicates both context building and information abstraction at different levels. Thus, naming and tagging conventions are needed, these conventions ought to be enforced using effective communication methods. 
\end{itemize}

Table \ref{tab:Challenges_Spec} summarizes the aforementioned challenges as well as others usually found in internal specifications that are expected to hinder the deployment of GenAI. It also includes their severity and potential actions to take in order to diminish their effect on model performance.

\section{Conclusions}
\label{sec:conclusion}

In this work, internal telecommunication specifications have been introduced and analyzed, also a vision of TLP for telecommunications has been built atop existing, general concepts. In this manner, TLP can help Telecommunications equipment vendors to improve information access, development, and engineer training times. State-of-the-art research focuses on the use of open-source technical specifications, as well as the training and evaluation of open-source models using this data for a myriad of use cases. By applying the TLP principles outlined in this work, vendors can start building their own internal tools that are adapted to their specific needs and documentation formats facilitating information extraction. 

As future research direction we would like to focus on the practical aspects of the issue including tokenization of technical text and covering the implementation and testing of a GenAI-based tool applying the principles exposed in this letter. Another aspect to consider is measuring the quality of the generated text, since state-of-the-art evaluation metrics are not applicable in this scenario.

\bibliographystyle{unsrtnat}
\bibliography{sample-base}  






\end{document}